\documentclass[letterpaper, 10 pt, conference]{ieeeconf}  % Comment this line out
\IEEEoverridecommandlockouts                              % This command is only
\usepackage[utf8]{inputenc}
\usepackage[T1]{fontenc}
\usepackage{graphicx}
\usepackage{amsfonts}
\usepackage[colorlinks,
            linkcolor = blue,
            anchorcolor = blue,
            urlcolor = blue,
            citecolor = blue,
            ]{hyperref}
\usepackage{float}%提供float浮动环境
\usepackage{booktabs}%提供命令\toprule、\midrule、\bottomrule
\usepackage{multirow}%提供跨列命令\multicolumn{}{}{}
\usepackage{cite}%文章引用
\usepackage{stfloats}%表格位置控制
\usepackage{amsmath}% 公式换行

\title{\LARGE \bf
CSFlow: Learning Optical Flow via Cross Strip Correlation for Autonomous Driving
}

\author{Hao Shi$^{1}$, Yifan Zhou$^{2}$, Kailun Yang$^{3}$, Xiaoting Yin$^{1}$, and Kaiwei Wang$^{1}$% <-this % stops a space
\thanks{*This work was partly supported by Data Intelligence Department of Luokung Technology Corp. (LKCO) and Sunny Optical Technology (Group) Co. Ltd. This work was also supported in part by the National Natural Science Foundation of China (NSFC) under Grant No. 12174341, in part by the Federal Ministry of Labor and Social Affairs (BMAS) through the AccessibleMaps project under Grant 01KM151112, in part by the University of Excellence through the ``KIT Future Fields'' project, and in part by Hangzhou SurImage Technology Co. Ltd.}% <-this % stops a space
\thanks{$^{1}$H. Shi, X. Yin, and K. Wang are with State Key Laboratory of Modern Optical Instrumentation, Zhejiang University, China {\tt\small \{haoshi, wangkaiwei, xiaotingyin\}@zju.edu.cn}}%
\thanks{$^{2}$Y. Zhou is with Shanghai Artificial Intelligence Laboratory, China {\tt\small zhouyifan@pjlab.org.cn}}%
\thanks{$^{3}$K. Yang is with Institute for Anthropomatics and Robotics, Karlsruhe Institute of Technology, Germany {\tt\small kailun.yang@kit.edu}}%
}

\begin{document}

\maketitle
\thispagestyle{empty}
\pagestyle{empty}

%%%%%%%%%%%%%%%%%%%%%%%%%%%%%%%%%%%%%%%%%%%%%%%%%%%%%%%%%%%%%%%%%%%%%%%%%%%%%%%%
\begin{abstract}
Optical flow estimation is an essential task in self-driving systems, which helps autonomous vehicles perceive temporal continuity information of surrounding scenes. The calculation of all-pair correlation plays an important role in many existing state-of-the-art optical flow estimation methods. However, the reliance on local knowledge often limits the model's accuracy under complex street scenes. In this paper, we propose a new deep network architecture for optical flow estimation in autonomous driving——\emph{CSFlow}, which consists of two novel modules: \emph{Cross Strip Correlation module (CSC)} and \emph{Correlation Regression Initialization module (CRI)}. CSC utilizes a striping operation across the target image and the attended image to encode global context into correlation volumes, while maintaining high efficiency. CRI is used to maximally exploit the global context for optical flow initialization. Our method has achieved state-of-the-art accuracy on the public autonomous driving dataset KITTI-2015. Code is publicly available at \url{https://github.com/MasterHow/CSFlow}.
\end{abstract}

%%%%%%%%%%%%%%%%%%%%%%%%%%%%%%%%%%%%%%%%%%%%%%%%%%%%%%%%%%%%%%%%%%%%%%%%%%%%%%%%
\section{INTRODUCTION}
\begin{figure}[!t]
   \setlength{\abovecaptionskip}{-0.2cm}   %调整图片标题与图距离
   \centering
   \includegraphics[scale=0.0564]{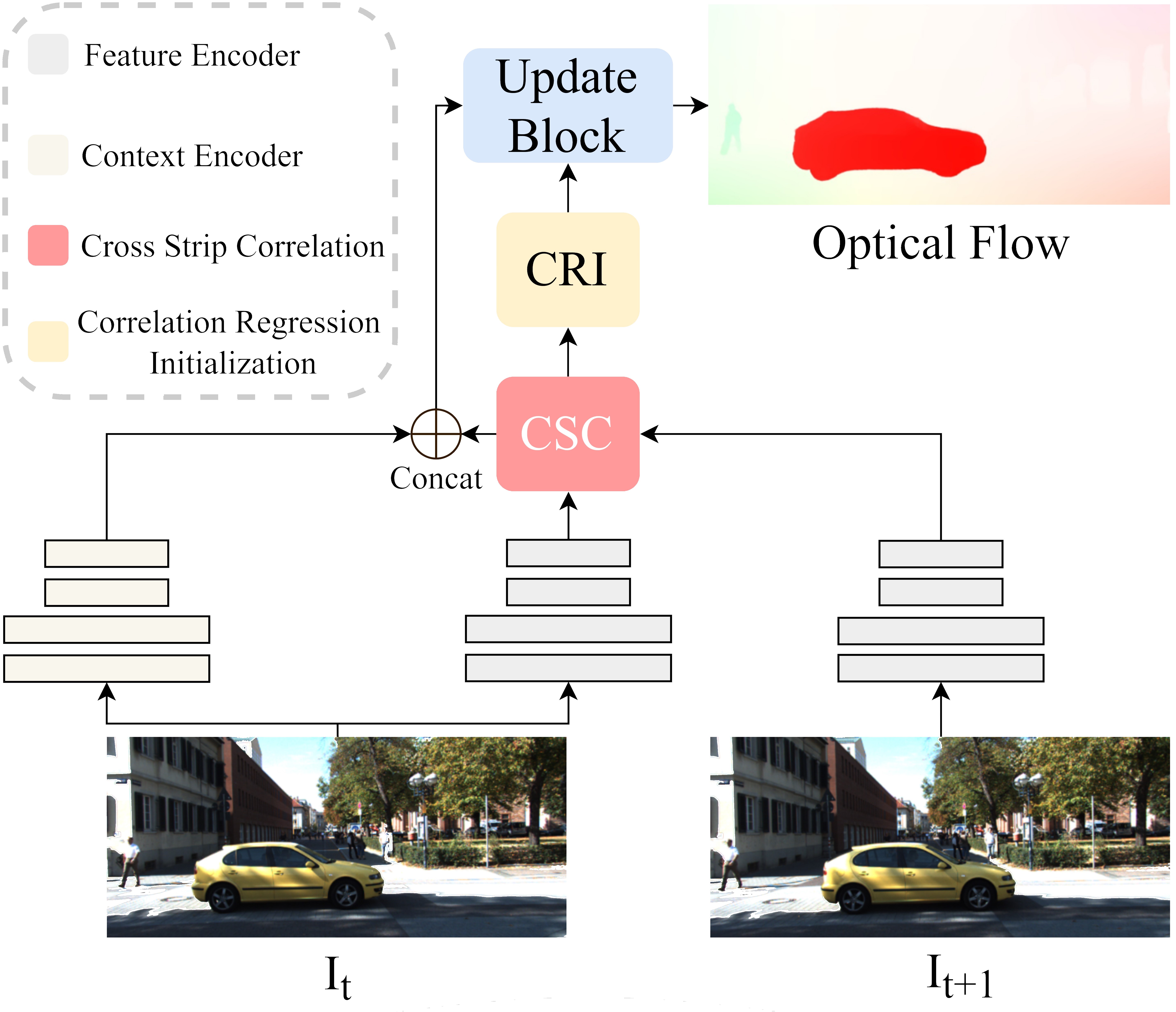} % 0.6 for png
   \caption{Overview of the proposed CSFlow architecture, which includes three main stages: 1) Feature extraction, 2) Cost volume calculation, and 3) Iterative refinement. We introduce Cross Strip Correlation module (CSC) to encode the global visual similarities into cost volume. The Correlation Regression Initialization module (CRI) will then exploit the orthogonal correlation to initialize the flow field without introducing any extra parameters, keeping the framework simple and efficient.}
   \label{fig:introduction_overview}
   \vspace{-1.75em}   %调整图片标题与下文距离
\end{figure}

Since Horn and Schunck’s pioneering work~\cite{horn1981determining}, the study of how to estimate per-pixel 2D motion between two time series images, \textit{i.e.}, optical flow estimation, has become a basic problem in computer vision.
It has been widely used in autonomous driving, 3D reconstruction, and robot sensing~\cite{kondermann2016hci,li2019learning,geiger2013vision}.
Specifically, flow estimation, as a fundamental task, helps vehicles perceive the temporal continuity of surrounding environmental features (see Fig.~\ref{fig:introduction_overview}), and thus it plays an important role in scene parsing, visual odometry, and SLAM systems~\cite{gadde2017semantic,mustikovela2016can,zhu2017deep,min2020voldor,teed2021droid}.
FlowNet~\cite{dosovitskiy2015flownet} utilizes Convolutional Neural Networks (CNNs) to estimate optical flow for the first time.
To improve the accuracy of the learning-based flow estimation, correlation volumes have been widely adopted in current popular optical flow frameworks~\cite{ilg2017flownet,hui2018liteflownet,sun2018pwc,teed2020raft}, proved to be able to effectively introduce the cost calculation of pixel matching into the model.
But in real applications, learning-based optical flow estimation still faces many challenges, including large displacements, motion blur, occlusion, and texture-less areas~\cite{teed2020raft,xu2021high}.
%图1 网络框架

In order to achieve fine-grained pixel matching, the current state-of-the-art method, RAFT~\cite{teed2020raft} uses all-pair correlation calculations and refines the optical flow through an iterative optimization framework.
However, all-pair correlation volume only includes the cost of local matching, and lacks awareness of the global context.
It is naive to rely solely on the iterative framework to deal with easily-confused areas.
High-level features are often computationally expensive to obtain, and non-local scene cues usually require heavy networks with large receptive fields to capture, such as ResNet-101~\cite{wang2018non} and DeepLab~\cite{chen2017deeplab}, which are impractical for optical flow estimation tasks.
In order to capture non-local context while maintaining high efficiency, we investigate the current commonly-used method, the self-attention mechanism~\cite{vaswani2017attention,woo2018cbam,cheng2016long}, which is capable of capturing long-distance dependencies.
However, it is observed that the calculation is still expensive, the computational complexity is usually $4$-dimensional \(O(H\times W\times H\times W)\) due to the need to calculate a huge attention map~\cite{wang2018non}.
Yet, we notice that optical flow can be naturally decomposed into two intersecting one-dimensional motion vectors, horizontally and vertically, which leads to an interesting rethinking perspective:
Is it possible to also decompose the global context into two cross directions and encode it in the correlation volume?

In this work, we propose a new model for real-world flow estimation based on the above observations——\emph{CSFlow}, which consists of two novel modules. \emph{Cross Strip Correlation module (CSC)} is motivated by the results of previous works~\cite{sun2018pwc,teed2020raft} when facing complex street scenes.
We find that various networks achieved low accuracy on thin and long strips such as poles and trees, and the error in texture-less areas can be further expanded.
It is reasonable as the convolution is a local operation and its receptive field is rather limited.
However, the previous cost volumes~\cite{yang2019volumetric,teed2020raft} cannot give accurate motion information when it lacks local direct matching information.
Therefore, we introduce strip operations in the traditional correlation calculations to increase directional consistency.
Inspired by the composition of optical flow, CSC also uses horizontal and vertical cross strip operations to propagate the global context in both directions. Specifically, cross striping operations encode non-local dependencies on both horizontal and vertical axes at the same time.
Each position in the target image is connected with pixels in different row and column spaces of the attended image, while the computational complexity is significantly reduced to \(O(H\times W\times (H+W))\).

Furthermore, we design a \emph{Correlation Regression Initialization module (CRI)} to maximize the use of cross strip correlation volumes and exploit global motion information. CRI does not contain any additional parameters, maintaining high efficiency with a minimal calculation.
The optical flow obtained through CRI regression can give motion information of large displacements at the initial stage of the iterative update, and propagate backward along the iteration to improve the optical flow refinement.

We conduct extensive quantitative experiments on popular optical flow datasets.
The synthetic to real generalization error of CSFlow on KITTI-2015 training set is reduced $14.1\%$ compared with RAFT (C→K, $32.3\%$ vs $37.6\%$), which is important for optical flow estimation, considering that it is difficult to obtain per-pixel optical flow ground truth in the real world.
CSFlow achieves an F1-all error of $5.00\%$ and an F1-noc error of $3.00\%$ on KITTI-2015 flow benchmark, ranked 2nd among all published results.
Moreover, we conduct ablation experiments, sufficiently demonstrating the effectiveness and performance of the proposed method.

In summary, our main contributions are as follows:
\begin{itemize}
   \item We have introduced a \emph{cross strip correlation module}, which can capture long-distance dependencies between two frames and encode the global context into the correlation calculation, while only slightly increasing the computational cost.
   \item We propose a \emph{correlation regression initialization module}, which can maximize the use of the non-local information given by the cross strip correlation to initialize the optical flow without any additional parameters.
   \item CSFlow not only achieves state-of-the-art performance on the KITTI-2015 flow benchmark, but also has better synthetic-to-real generalization capabilities than previous flow estimation methods.
\end{itemize}

\section{RELATED WORK}
Optical flow estimation has been traditionally regarded as an optimization problem. Since the introduction of FlowNet~\cite{dosovitskiy2015flownet} in 2015, the learning-based optical flow estimation has gradually surpassed the traditional methods represented by the variational method in both accuracy and speed~\cite{ilg2017flownet,hui2018liteflownet,hui2020lightweight}. In this section, we focus on reviewing learning-based methods from different perspectives.
\subsection{End-To-End Flow Prediction}
Dosovitskiy~\textit{et al.}~\cite{dosovitskiy2015flownet} use Convolutional Neural Networks (CNNs) to estimate optical flow for the first time.
In order to reduce computational complexity and memory requirements, Sun~\textit{et al.}~\cite{sun2018pwc} construct a partial cost volume at multiple pyramid levels.
However, such coarse-to-fine methods tend to miss fast-moving small objects~\cite{revaud2015epicflow} when the resolution is too coarse.
To alleviate this issue, RAFT~\cite{teed2020raft} proposes to maintain a single high-resolution feature and refine the initial prediction with a large number of iterative refinements, achieving state-of-the-art performance on standard benchmarks.
Following the basic framework of RAFT, we use convolutional GRU units to update a fixed high-resolution optical flow field. Instead of initializing the optical flow to zeros, we obtain the flow distribution by regressing the cross strip correlation before the first iteration, which helps the update block focus on the areas that are easily confused.
\subsection{Cost Volumes}
Cost volume is a concept that often appears in Deep Neural Networks (DNNs) for optical flow~\cite{hofinger2020improving,hosni2012fast,hui2020liteflownet3,lu2020devon,sun2018pwc,teed2020raft,wang2020displacement,yang2019volumetric}, which stores the matching costs for each pixel’s potential correspondences.
It serves as a discriminative representation of the search space and usually leads to accurate results by utilizing powerful optimization methods~\cite{hirschmuller2007stereo,hosni2012fast}.
The current state-of-the-art optical flow model, RAFT~\cite{teed2020raft} builds multi-scale 4D cost volumes by computing all spatial correlations.
However, the all-pair correlation in RAFT only encodes the per-pixel dense matching cost. This leads to areas with similar texture features of the foreground and background being easily confused, resulting in estimation failures.
Inspired by the fact that optical flow can be naturally decomposed into orthogonal directions, we calculate the correlation volumes by associating the target image and the feature map of the attended image in both horizontal and vertical directions, introducing non-local scene cues into the cost volume.
Recently, Flow1D~\cite{xu2021high} also starts from the decomposition of correlation, using 1D attention and 1D correlation to obtain cost volume.
Instead of using computationally expensive attention mechanisms to generate key values, we introduce strip operation to directly obtain orthogonal energy maps from the activated feature maps.
This allows to maintain high efficiency when encoding the global context, and benefits from all-pair correlation to transmit the fine-grained matching cost at the same time, remaining high accuracy for complex street scene understanding.
\section{APPROACH}
\begin{figure*}[!t]
   \centering
   \includegraphics[scale=0.75]{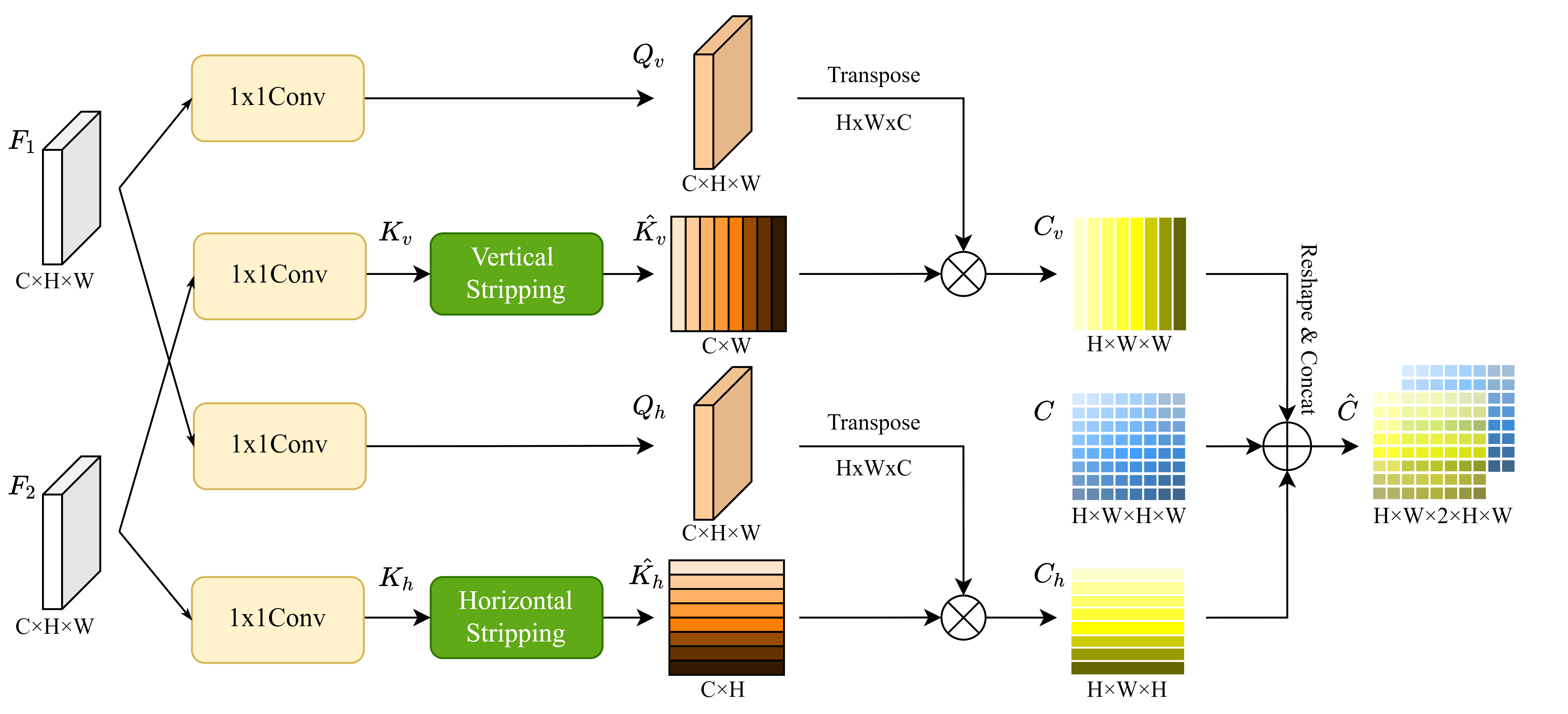}
   \caption{\emph{Cross Strip Correlation module}. Orthogonal queries are extracted from the feature map of the target image. The vertical and horizontal stripping are introduced to aggregate non-local scene clues in the attended image. The orthogonal cost volume can then be obtained via a cross correlation of query and key maps, giving the per-pixel matching cost. We benefit from the inexpensive computing needs of strip operation, which allows us to maintain the fine-grained all-pair correlation at the same time.}
   \label{fig:cross_strip_correlation}
   \vspace{-1.5em}   %调整图片标题与下文距离
\end{figure*}
\begin{figure}[h]
   \centering
   \includegraphics[scale=0.085]{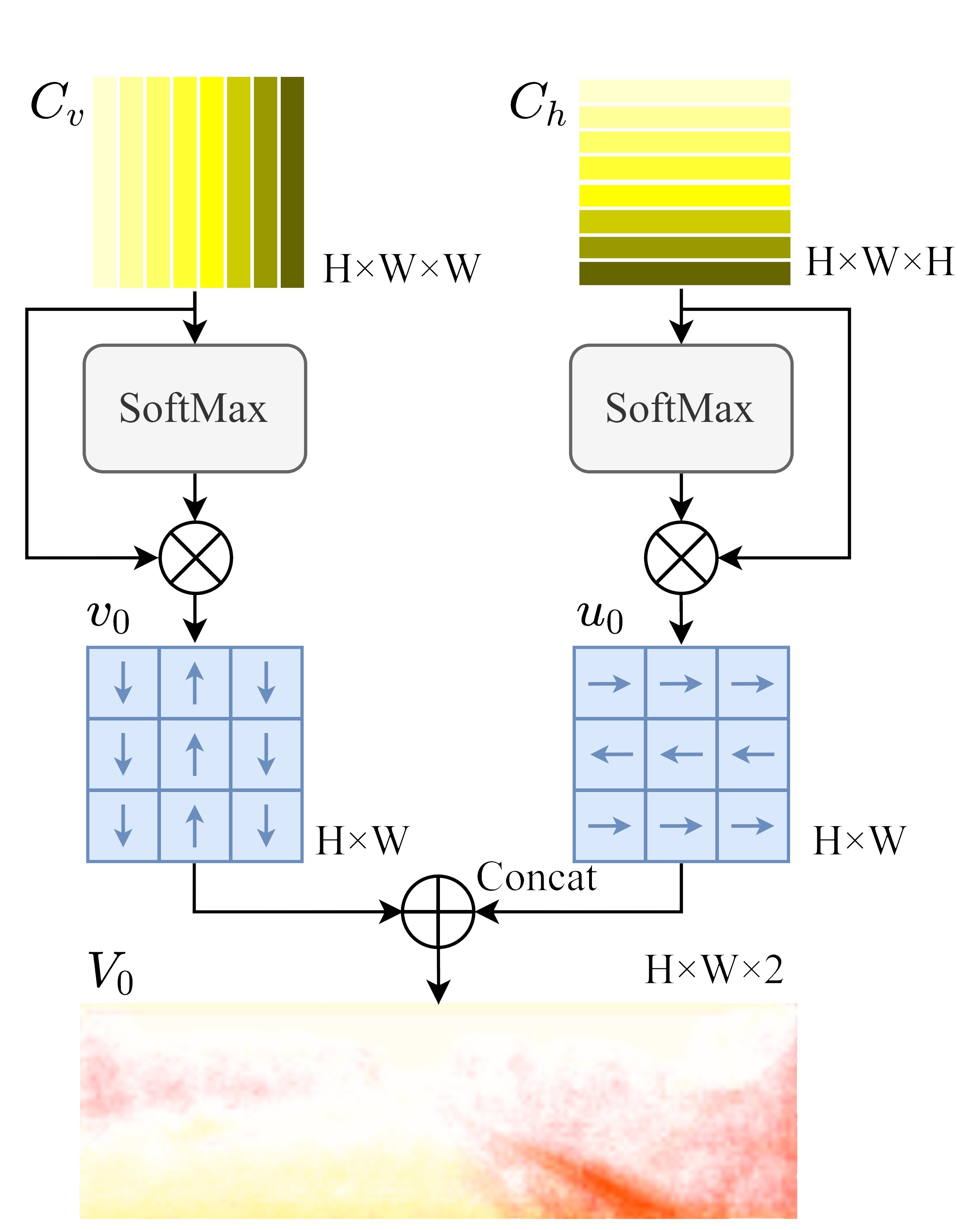} % 0.9 for png
   \caption{\emph{Correlation Regression Initialization module}. The initial flow field can be obtained by regressing the orthogonal correlation in two directions separately, and all the operations are parameters-free.}
   \label{fig:correlation_regression_initialization}
   \vspace{-1.0em}   %调整图片标题与下文距离
\end{figure}

We first describe the basic architecture of CSFlow. In order to capture long-distance dependencies, we introduce a \emph{Cross Strip Correlation module (CSC)} (Fig.~\ref{fig:cross_strip_correlation}) to calculate the dense matching cost of optical flow, and introduce a \emph{Correlation Regression Initialization module (CRI)} (Fig.~\ref{fig:correlation_regression_initialization}) to further efficiently use the global context.

\subsection{Network Architecture}
The structure of CSFlow is shown in Fig.~\ref{fig:introduction_overview}.
We use RAFT~\cite{teed2020raft} as a baseline to verify the effectiveness of our proposed orthogonal correlation method.
Given two frames of sequence RGB images $I_1$ and $I_2$, we estimate the dense motion vector $(u, v)$ from each pixel $(x, y)$ of $I_1$ to each pixel $(x', y')$ of $I_2$, that is, the optical flow field $V$, which gives the per-pixel mapping relationship between the attended image and the target image. Our method can be divided into the following three stages: feature extraction, calculation of cost volume, and iterative refinement.

\subsubsection{Feature Extraction.}
We use a convolutional network as the encoder $e(\cdot)$ to extract features at $1/8$ resolution $F_1, F_2 \in \mathbb{R} ^ {C \times H \times W}$ from the input two frames of images $I_1$ and $I_2$. The weights of the encoders on the two branches are not shared. A context encoder $c(\cdot)$ is also introduced with the same structure as the feature encoder to extract context information from $I_1$. The context is then concatenated with the aggregated correlation value $\hat{C}$ given by the CSC module and sent to the update block to iteratively update the fixed resolution flow field.

\subsubsection{Cost Volume Calculation.}
The CSC module is introduced to calculate the orthogonal correlation, and then we concatenate it with the all-pair correlation, which is used for the initialization of the flow field and guides the refinement of the optical flow. The all-pair correlation $C$ can be directly obtained by calculating the tensor dot product of $F_1$ and $F_2$ obtained by feature extraction of $I_1$ and $I_2$, as depicted in Equ.~(\ref{equ:dot_product}).
The details of the CSC module will be described in the next subsection.
\begin{equation}
\label{equ:dot_product}
   C(x_1, y_1, x_2, y_2) = F_1(x_1, y_1)^T \cdot  F_2(x_2, y_2)
\end{equation}

\subsubsection{Iterative Refinement.}
Following previous works\cite{teed2020raft,xu2021high}, we use the convolutional GRU unit as the update block to iteratively refine the flow field. The input of the update block includes the context information $C_1$ obtained by $I_1$ through the context encoder $c(\cdot)$, the aggregated correlation volume $\hat{C}$ given by the CSC module, and the initialization flow provided by the CRI module.
The update block will estimate an optical flow sequence $\{V_1, V_2, {...} V_m\}$, where $m$ is set as $12$ during training.
The optical flow $V_0$ is obtained by regression of the orthogonal correlation through the CRI module.
The details of the CRI module will be described below.
Each iteration will produce an update field ${\Delta}V$, thus the updated optical flow is $V_{i+1}=V_i+{\Delta}V$.
After each update, $V_i$ will be upsampled to the original input scale.
We upsample the optical flow to full resolution by taking the convex combination of the low-resolution $3 \times 3$ grid.
After that, we calculate the L1 norm between the predicted $V_i$ and the ground truth $V_t$ with the supervision loss function, and the initial estimate $V_0$ given by CRI is also supervised.
For the estimation of flow sequence $\{V_0, V_1, V_2$, ... $V_m \}$, we calculate the weighted sequence loss function:
\begin{equation}
\label{equ:weighted_sequence_loss}
   \mathcal{L}  = \sum_{i = 0}^{m} \Upsilon ^ {N - i} \Vert V_t - V_i \Vert_{1}.
\end{equation}
In the pre-training stage, $\Upsilon$ is set to $0.8$, and in the fine-tuning stage, the value of $\Upsilon$ is adjusted to $0.85$.

\subsection{Cross Strip Correlation Module}
To capture the global context and reduce the computational complexity of non-local information encoding, we introduce a \emph{Cross Strip Correlation module (CSC)}. The structure of the module is shown in Fig.~\ref{fig:cross_strip_correlation}.
CSC extracts the orthogonal query matrix $Q_v$ and $Q_h$ of the target image and the orthogonal key matrix $\hat{K_v}$ and $\hat{K_h}$ of the involved images through the strip operation, while keeping the direction consistency of the global information.
Subsequently, we further calculate the correlation between the orthogonal query matrix and the orthogonal key matrix to obtain the global visual similarity of $I_1$ and $I_2$ in the vertical and horizontal directions to guide the iterative updates of flow field.

Specifically, given $I_1$, $I_2$, the feature map $F_1, F_2 \in \mathbb{R}  ^ {C \times H \times W}$ are obtained by the feature encoder $e()$. We first introduce two $1 \times 1$ convolutional layers to activate $F_1$ and obtain vertical and horizontal query matrix $Q_v, Q_h \in C^{'} \times H \times W$. Simultaneously, we use two additional $1 \times 1$ convolutional layers to activate $F_2$ and obtain $K_v, K_h \in C^{'} \times H \times W$. Then, the orthogonal global key matrix $\hat{K_v}\in C^{'} \times W, \hat{ K_h} \in C^{'} \times H $ can be obtained through vertical and horizontal striping operations:
\begin{equation}
   \begin{split}
      \left\{
      \begin{aligned}
         \hat{K_v}(i, j) = \frac{1}{H}\sum_{k = 1}^{H} K_v(i, k, j), \\
         \hat{K_h}(i, j) = \frac{1}{W}\sum_{k = 1}^{W} K_h(i, j, k),
      \end{aligned}
      \right.
   \end{split}
\end{equation}
where the average pooling window is $H \times 1$ and $1 \times W$, respectively. The intuition of the operation is that it emphasizes the vertical- and horizontal feature, which corresponds to the definition of optical flow. Then we transpose the orthogonal query matrix from $C^{'} \times H \times W$ to $H \times W \times C^{'}$, and perform the dot product with the orthogonal key values to obtain the vertical and horizontal correlation volumes $C_v \in H\times W \times W, C_h \in H \times W \times H$, which encode the non-local visual similarity:
\begin{equation}
   \begin{split}
      \left\{
      \begin{aligned}
         C_v(x) & = & Q_v(x) \cdot \hat{K_v}, \\
         C_h(x) & = & Q_h(x) \cdot \hat{K_h}.
      \end{aligned}
      \right.
   \end{split}
\end{equation}
$C_v$ and $C_h$ are then concatenated with the all-pair correlation volume $C\in H \times W \times H \times W $ to obtain an aggregated correlation $\hat{C} \in H \times W \times 2 \times H \times W$.
$\hat{C}$ encodes both fine-grained matching costs and global visual similarity. $\hat{C}$ is then sent to a $4$-layer average pooling pyramid with kernel sizes $1, 2, 4, 8$ which pools the last two dimensions to establish a $4$-layer correlation pyramid {$\hat{C_1}$, $\hat{C_2}$, $\hat{C_3}$, $\hat{C_4}$}.
Since the high-resolution information of the first two dimensions are preserved, the correlation pyramid can give the displacement information of fast-moving small objects.
Additionally, the correlation pyramid further aggregates non-local scene cues, assisting the update block to distinguish the optical flow of extreme displacements on similar textures.

\subsection{Correlation Regression Initialization Module}
As mentioned above, we introduce the \emph{Correlation Regression Initialization (CRI) module} to fully take the advantages of the orthogonal correlations $C_v$, $C_h$ given by CSC.
We regress the orthogonal correlation volumes instead of the all-pair correlation $C$.
The reason is that $C$ contains redundant spatial details, which introduce useless information during regressing.
We will verify this in the ablation experiments ( Sec.~\ref{sec:experiments}).
The rich high-level context information in $C_v$ and $C_h$ are more helpful to obtain the initialized optical flow field.
In consideration of the speed performance important for autonomous driving, we use the CRI module without any learnable parameters to regress $C_v$ and $C_h$.

The details of the CRI module are shown in Fig.~\ref{fig:correlation_regression_initialization}.
Given the orthogonal correlation volumes
$C_v \in H \times W \times W, C_h \in H \times W \times H$, we use a softmax layer to activate $C_v$ and $C_h$ respectively, and perform the multiplication on each element of original $C_v$ and $C_h$ to obtain the orthogonal energy maps.
Finally, we can obtain the vertical and horizontal initialization of optical flow $v_0, u_0 \in H \times W$:
\begin{equation}
\begin{split}
      \left\{
      \begin{aligned}
         v_0(x, y) & = & \sum_{w = 1}^{W}  \sigma(C_v(x, y, w))C_v(x, y, w), \\
         h_0(x, y) & = & \sum_{h = 1}^{H}  \sigma(C_h(x, y, h))C_h(x, y, h).
      \end{aligned}
      \right.
   \end{split}
\end{equation}

We then stack the orthogonal optical flows $v_0$ and $v_1$ to attain the initialized flow field $V_0 \in H \times W \times 2$, which will be further sent to the update block as the initial distribution of optical flow refinement. The module takes the advantage of global visual similarity to initialize the optical flow, lowering the workload of the subsequent iterative updates, so that the update block can focus on distinguishing foreground and background and estimating the motion information of the area with similar textures.

\section{EXPERIMENTS}
\label{sec:experiments}
We evaluate CSFlow on Sintel~\cite{butler2012naturalistic} and KITTI~\cite{geiger2013vision} datasets.
The model is first pre-trained on the synthetic datasets FlyingChairs~\cite{dosovitskiy2015flownet} and FlyingThings~\cite{mayer2016large}, and fine-tuned on Sintel or KITTI-2015. In the following, we present the experiment settings and results.

\subsection{Training Details}
CSFlow is trained on a NVIDIA Tesla P40, implemented in PyTorch.
All the weights are initialized randomly.
We choose AdamW optimizer~\cite{loshchilov2017decoupled}, and set the learning rate according to the one-cycle learning rate policy~\cite{smith2019super}. During the training process, the flow field is updated $12$ times.
When evaluating on KITTI, we set the number of updates to $24$.
When evaluating on Sintel, the number of updates is set to $32$.
The final model is first trained on FlyingChairs (C) with $150k$ iterations, where the batch size is $10$, the learning rate is $4e^{-4}$, and the image size is cropped to $368{\times}496$. Then $150k$ training iterations are performed on FlyingThings (T), where the batch size is $6$, the learning rate is $1.25e^{-4}$, and the image size is $400{\times}720$. Finally, we fine-tune the model on Sintel (S) or KITTI-2015 (K). The ablation experiment is performed with $100k$ training iterations on synthetic data Chairs, and the batch size is $10$.
When evaluating the results of ablation experiments, the flow field is updated $32$ times.
Following the settings of RAFT, all experiments are with data augmentation, including spatial and photometric augmentation.

\subsection{Zero-Shot Generalization}
We evaluate the synthetic-to-real zero-shot generalization capability of CSFlow on KITTI-12 and KITTI-15. Our model is trained with $150k$ iterations on Chairs (C) and Things (T). The model is trained on synthetic data and then directly evaluated on unseen real-world datasets. We compare the previous methods under the same zero-shot setting. As shown in Tab.~\ref{tab:generalization}, CSFlow exhibits state-of-the-art performance under complex street scenes in the zero-shot setting. When trained only on Chairs (C), CSFlow achieves an F1-all error of $22.3\%$ on KITTI-12, a $27.1\%$ error reduction compared with RAFT. After the C+T training, CSFlow achieves an end-point-error of $1.96$ pixels on KITTI-12 and an $8.4\%$ error reduction from the best prior network trained on the same data ($2.14$ pixels). This feature is vital in autonomous driving applications, considering that there is no large-scale real-world optical flow data available for training. 
%泛化表 
\begin{table}[!t]
\scriptsize
   %\vspace{-1.5em}
   \caption{\textbf{Synthetic to real generalization experiments.}}
   \label{tab:generalization}
   \centering
   \renewcommand\arraystretch{1.5}{\setlength{\tabcolsep}{1.5mm}{\begin{tabular}{cccccc}%四个c代表有四列且内容居中
            \toprule%第1道横线
            \multirow{2}{*}{Training data} & \multirow{2}{*}{Method}               & \multicolumn{2}{c}{\underline{KITTI-12 (train)}} & \multicolumn{2}{c}{\underline{KITTI-15 (train)}}                                       \\
                                           &                                       & F1-epe                                           & F1-all                                           & F1-epe           & F1-all           \\
            \midrule%第2道横线
            \multirow{4}{*}{C}             & PWC-Net\cite{sun2018pwc}              & 5.14                                             & \underline{28.7}                                 & 13.2             & 41.8             \\
                                           & RAFT\cite{teed2020raft}               & 4.72                                             & 30.6                                             & \underline{9.86} & 37.6             \\
                                           & LiteFlowNet2\cite{hui2020lightweight} & \underline{4.11}                                 & -                                                & 11.31            & \textbf{32.1}    \\
                                           & Ours                                  & \textbf{4.05}                                    & \textbf{22.3}                                    & \textbf{9.28}    & \underline{32.3} \\
            \midrule%第3道横线
            \multirow{9}{*}{C+T}           & PWC-Net                               & 4.14                                             & 21.4                                             & 10.35            & 33.7             \\
                                           & FlowNet2\cite{ilg2017flownet}         & 4.09                                             & -                                                & 10.06            & 30.0             \\
                                           & LiteFlowNet2\cite{hui2020lightweight} & 3.42                                             & -                                                & 8.97             & 25.9             \\
                                           & VCN\cite{yang2019volumetric}          & -                                                & -                                                & 8.36             & 25.1             \\
                                           & HD3\cite{yin2019hierarchical}         & 4.65                                             & -                                                & 13.17            & 24.0             \\
                                           & DICL-Flow\cite{wang2020displacement}  & -                                                & -                                                & 8.70             & 23.6             \\
                                           & MaskFlowNet\cite{zhao2020maskflownet} & 2.94                                             & -                                                & -                & 23.1             \\
                                           & RAFT\cite{teed2020raft}               & \underline{2.14}                                 & \underline{9.3}                                  & \underline{5.04} & \underline{17.4} \\
                                           & Ours                                  & \textbf{1.96}                                    & \textbf{8.6}                                     & \textbf{4.69}    & \textbf{16.5}    \\
            \bottomrule%第4道横线
         \end{tabular}}}
         \vspace{-0.8em}
\end{table}

\subsection{Ablations}
\begin{table*}[!t]
   \caption{\textbf{Ablation experiment results.}}
   \label{tab:ablation}
   \centering
   \renewcommand\arraystretch{1.4}{\setlength{\tabcolsep}{4.73mm}{\begin{tabular}{ccccccccc}%四个c代表有四列且内容居中
            \toprule%第1道横线
            \multirow{2}{*}{Experiment}             & \multirow{2}{*}{Variations} & \multicolumn{2}{c}{\underline{Sintel (train)}} & \multicolumn{2}{c}{\underline{KITTI-12 (train)}} & \multicolumn{2}{c}{\underline{KITTI-15 (train)}} & \multirow{2}{*}{Params.}                           \\%Data1,Data2跨两行，自动表格宽度
                                                    &                             & Clean                                          & Final                                            & F1-epe                                           & F1-all                   & F1-epe & F1-all         \\
            \midrule%第2道横线 Baseline
            Baseline\cite{teed2020raft}             & -                           & 2.26                                           & 4.50                                             & 4.72                                             & 30.6                     & 9.86   & 37.6   & 5.3M  \\%Data1跨两行，自动表格宽度
            \midrule%第8道横线 SCv2
            \multirow{2}{*}{CRI}                    & w/o                         & 2.28                                           & 4.35                                             & 4.85                                             & 26.3                     & 10.88  & 35.3   & 5.6M  \\%Data1跨两行，自动表格宽度
                                                    & \underline{with}            & 2.17                                           & 3.98                                             & 4.05                                             & 22.3                     & 9.28   & 32.3   & 5.6M  \\ %SCv2+CIv1
            \midrule%第5道横线 SCv2+CIv3
            \multirow{2}{*}{Initial method}         & Flow head                   & 2.25                                           & 4.03                                             & 4.63                                             & 26.0                     & 10.04  & 35.7   & 6.9M  \\%Data1跨两行，自动表格宽度
                                                    & \underline{Regression}      & 2.17                                           & 3.98                                             & 4.05                                             & 22.3                     & 9.28   & 32.3   & 5.6M  \\ %SCv2+CIv1
            \midrule%第4道横线 SCv2+CIv2
            \multirow{2}{*}{Correlation Regression} & All-pairs                   & 2.48                                           & 4.16                                             & 4.83                                             & 26.7                     & 11.32  & 37.3   & 5.6M  \\%Data1跨两行，自动表格宽度
                                                    & \underline{Strip-corrs}     & 2.17                                           & 3.98                                             & 4.05                                             & 22.3                     & 9.28   & 32.3   & 5.6M  \\ %SCv2+CIv1
            \midrule%第6道横线 SCv2+CIv1+MG
            \multirow{2}{*}{GRU levels}             & 3                           & 2.46                                           & 4.19                                             & 4.45                                             & 24.1                     & 10.54  & 34.7   & 12.3M \\%Data1跨两行，自动表格宽度
                                                    & \underline{1}               & 2.17                                           & 3.98                                             & 4.05                                             & 22.3                     & 9.28   & 32.3   & 5.6M  \\ %SCv2+CIv1
            \midrule%第7道横线 CIv2
            \multirow{2}{*}{CSC}                    & w/o                         & 2.59                                           & 4.52                                             & 6.49                                             & 41.6                     & 13.12  & 46.4   & 5.3M  \\
                                                    & \underline{with}            & 2.17                                           & 3.98                                             & 4.05                                             & 22.3                     & 9.28   & 32.3   & 5.6M  \\ %SCv2+CIv1
            \midrule%第3道横线 SCv1+CIv1
            \multirow{2}{*}{Queries}                & Same                        & 2.37                                           & 5.00                                             & 4.91                                             & 28.5                     & 10.82  & 37.6   & 5.5M  \\%Data1跨两行，自动表格宽度
                                                    & \underline{Separate}        & 2.17                                           & 3.98                                             & 4.05                                             & 22.3                     & 9.28   & 32.3   & 5.6M  \\ %SCv2+CIv1
            \bottomrule%第9道横线
         \end{tabular}}}
         \vspace{-1.0em} %减少表格与下文间距
\end{table*}

\begin{figure*}[t]
   \centering
   \includegraphics[scale=0.83]{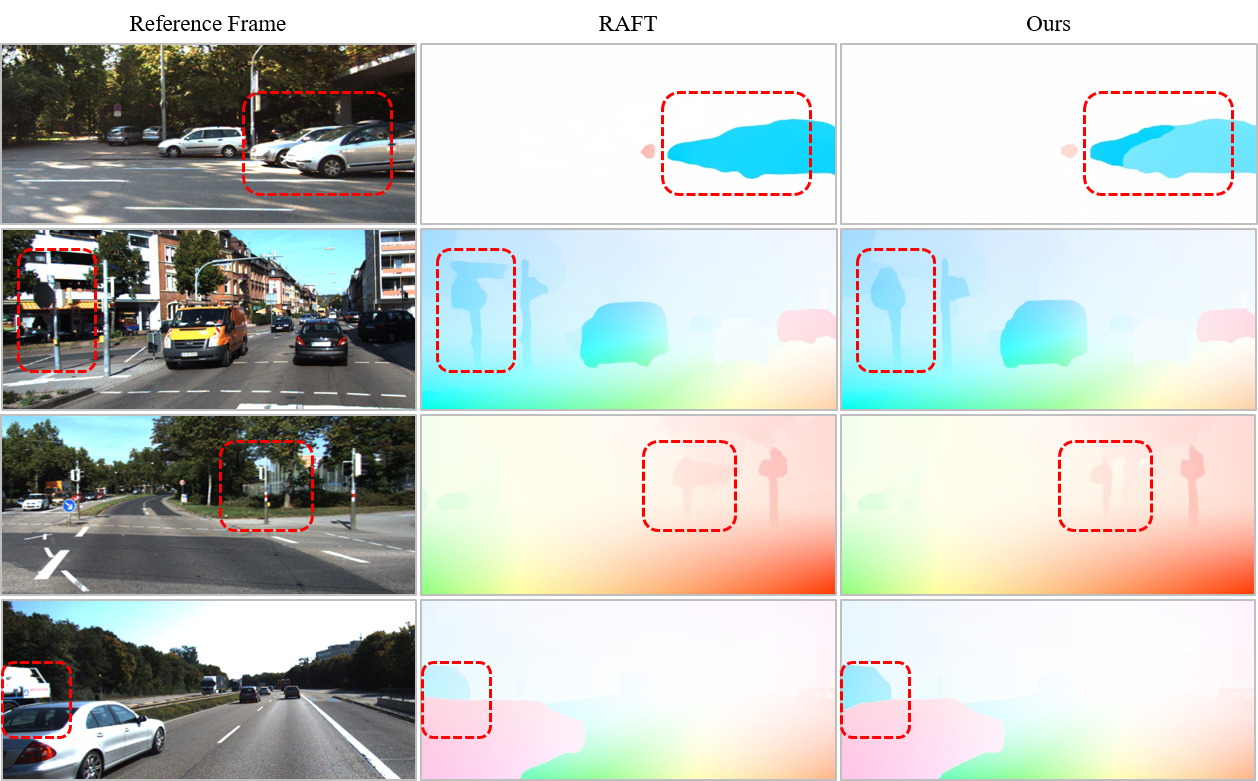}
   \caption{Qualitative comparisons on KITTI-2015 test split. We compare the proposed CSFlow with the current state-of-the-art method~\cite{teed2020raft}. Both models are fine-tuned on KITTI-2015 training set. Significant improvements are highlighted by rectangle. Due to the directional non-local context given by cross strip correlation, our method can overcome easily-confused situations such as texture-less or similar texture areas. Best viewed in color.}
   \label{fig:qualitative}
   \vspace{-1.5em} %减少表格与下文间距
\end{figure*}

We conduct a series of ablation experiments to verify the component requirements of the proposed CSFlow. As shown in Tab.~\ref{tab:ablation}, the settings used in the final model are underlined. We test each component individually, while keeping the settings of other components consistent with the final model, and evaluate the model on Sintel and KITTI.

\textit{CRI:}
The optical flow initialization method we propose significantly improves the estimation accuracy without additional parameters.
\textit{Initial method:} Flow head uses an additional convolutional layer to process the correlation volume and calculate the initial optical flow, which improves the result compared to the baseline. However, we notice that it is more efficient to initialize the optical flow using regression without introducing additional calculations.
\textit{Correlation Regression:}
Compared to directly regressing all pairs correlation volume, it is better to use orthogonal correlations to initialize optical flow, because it makes greater use of non-local visual similarity.
\textit{GRU Levels:}
Following RAFT-Stereo~\cite{lipson2021raft}, we adapt the multi-layer GRU from the parallax estimation task to optical flow. Multi-layer GRU maintains and updates $1/32$, $1/16$, and $1/8$ multi-scale optical flow fields, respectively. However, we find that this change introduces a large number of parameters, while the improvement is negligible.
\textit{CSC:}
Our proposed cross strip correlation aggregates the global context, leading to an end-point-error of 3.98 pixels on Sintel-final, an $11.6\%$ error reduction \textit{w.r.t.} the baseline.
\textit{Queries:} We further try to use the same query volume for $I_1$ to calculate the orthogonal correlation from $I_2$, which slightly reduces the amount of parameters. Using the separate query volume significantly improves the estimation accuracy, which shows that two orthogonal query volumes are required for the horizontal and vertical key matrices.

\subsection{Quantitative Experiment}
\begin{table*}[t]
   \caption{\textbf{Quantitative results on Sintel and KITTI 2015 datasets.}}
   \label{tab:quantitative}
   \centering
   \renewcommand\arraystretch{1.5}{\setlength{\tabcolsep}{4.35mm}{\begin{tabular}{cccccccccc}%四个c代表有四列且内容居中，1.5好看但是放不下。
            \toprule%第1道横线
            \multirow{2}{*}{Training data} & \multirow{2}{*}{Method}               & \multicolumn{2}{c}{\underline{Sintel (train)}} & \multicolumn{2}{c}{\underline{KITTI-15 (train)}} & \multicolumn{2}{c}{\underline{Sintel (test)}} & \multicolumn{2}{c}{\underline{KITTI-15 (test)}}                                                                             \\
                                           &                                       & Clean                                          & Final                                            & F1-epe                                        & F1-all                                          & Clean            & Final            & F1-fg            & F1-all           \\
            \midrule%第2道横线
            \multirow{9}{*}{C+T}           & PWC-Net\cite{sun2018pwc}              & 2.55                                           & 3.93                                             & 10.35                                         & 33.7                                            & -                & -                & -                & -                \\
                                           & FlowNet2\cite{ilg2017flownet}         & 2.02                                           & 3.54                                             & 10.08                                         & 30.0                                            & 3.96             & 6.02             & -                & -                \\
                                           & VCN\cite{yang2019volumetric}          & 2.21                                           & 3.68                                             & 8.36                                          & 25.1                                            & -                & -                & -                & -                \\
                                           & HD3\cite{yin2019hierarchical}         & 3.84                                           & 8.77                                             & 13.17                                         & 24.0                                            & -                & -                & -                & -                \\
                                           & DICL-Flow\cite{wang2020displacement}  & 1.94                                           & 3.77                                             & 8.70                                          & 23.6                                            & -                & -                & -                & -                \\
                                           & MaskFlowNet\cite{zhao2020maskflownet} & 2.25                                           & 3.61                                             & -                                             & 23.1                                            & -                & -                & -                & -                \\
                                           & Flow1D\cite{xu2021high}               & 1.98                                           & 3.27                                             & 6.69                                          & 23.0                                            & -                & -                & -                & -                \\
                                           & RAFT\cite{teed2020raft}               & \underline{1.43}                               & \underline{2.71}                                 & \underline{5.04}                              & \underline{17.4}                                & -                & -                & -                & -                \\
                                           & Ours                                  & \textbf{1.42}                                  & \textbf{2.60}                                    & \textbf{4.69}                                 & \textbf{16.5}                                   & -                & -                & -                & -                \\
            \midrule%第3道横线
            \multirow{9}{*}{C+T+S/K}       & FlowNet2\cite{ilg2017flownet}         & (1.45)                                         & (2.01)                                           & (2.30)                                        & (8.61)                                          & 4.16             & 5.74             & 8.75             & 10.41            \\
                                           & PWC-Net\cite{sun2018pwc}              & -                                              & -                                                & -                                             & -                                               & 4.39             & 5.04             & 9.31             & 9.60             \\
                                           & IRR-PWC\cite{hur2019iterative}        & (1.92)                                         & (2.51)                                           & (1.63)                                        & (5.32)                                          & 3.84             & 4.58             & 7.52             & 7.65             \\
                                           & ScopeFlow\cite{bar2020scopeflow}      & -                                              & -                                                & -                                             & -                                               & 3.59             & 4.10             & \underline{7.36} & 6.82             \\
                                           & HD3\cite{yin2019hierarchical}         & (1.87)                                         & (1.17)                                           & (1.31)                                        & (4.10)                                          & 4.79             & 4.67             & 9.02             & 6.55             \\
                                           & Flow1D\cite{xu2021high}               & (0.84)                                         & (1.25)                                           & -                                             & (1.60)                                          & 2.24             & 3.81             & -                & 6.27             \\
                                           & RAFT+LCV\cite{xiao2020learnable}      & (0.94)                                         & (1.31)                                           & (1.06)                                        & (3.77)                                          & 2.75             & 3.55             & 8.90             & 6.26             \\
                                           & RAFT\cite{teed2020raft}               & (0.77)                                         & (1.20)                                           & (0.64)                                        & (1.50)                                          & \underline{2.08} & \underline{3.41} & -                & \underline{5.27} \\
                                           & Ours                                  & (0.64)                                         & (1.01)                                           & (0.80)                                        & (2.14)                                          & \textbf{1.84}    & \textbf{3.37}    & \textbf{6.10}    & \textbf{5.06}    \\
            \midrule%第4道横线
            \multirow{7}{*}{C+T+S+K+H}     & PWC-Net+\cite{sun2019models}          & (1.71)                                         & (2.34)                                           & (1.50)                                        & (5.3)                                           & 3.45             & 4.60             & 7.88             & 7.72             \\
                                           & LiteFlowNet2\cite{hui2020lightweight} & (1.30)                                         & (1.62)                                           & (1.47)                                        & (4.8)                                           & 3.48             & 4.69             & 7.64             & 7.62             \\
                                           & VCN\cite{yang2019volumetric}          & (1.66)                                         & (2.24)                                           & (1.16)                                        & (4.1)                                           & 2.81             & 4.40             & 8.66             & 6.30             \\
                                           & MaskFlowNet\cite{zhao2020maskflownet} & -                                              & -                                                & -                                             & -                                               & 2.52             & 4.17             & 7.70             & 6.11             \\
                                           & RAFT (2-view)\cite{teed2020raft}      & (0.76)                                         & (1.22)                                           & (0.63)                                        & (1.50)                                          & \underline{1.94} & \underline{3.18} & \underline{6.87} & \underline{5.10} \\
                                           & Ours (2-view)                         & (0.62)                                         & (0.98)                                           & (0.71)                                        & (1.76)                                          & \textbf{1.63}    & \textbf{3.03}    & \textbf{6.46}    & \textbf{5.00}    \\
            \cline{2-10}%间隔横线
                                           & RAFT†\cite{teed2020raft}              & (0.77)                                         & (1.27)                                           & -                                             & -                                               & 1.61             & 2.86             & -                & -                \\
            \bottomrule%第6道横线
         \end{tabular}}}
\end{table*}
We quantitatively evaluate CSFlow on Sintel and KITTI-2015. As shown in Tab.~\ref{tab:quantitative}, the best results are shown in bold, and the second best results are underlined, † indicates that the method uses the estimation result of the previous frame to refine the subsequent optical flow. The model has been trained for two stages. The first stage is pre-training on synthetic data Chairs (C) and Things (T), and the second stage is fine-tuning on Sintel (S) and KITTI (K) respectively. Following RAFT, our final model is divided into two types, one is to fine-tune only on the training set of the target benchmark (C+T+S/K), while the other is to fine-tune the mixed data (C+T+S+K+H). Our mixed data distribution is consistent with RAFT. We use average End Point Error (EPE) and percentage of optical flow outliers (F1) as evaluation indicators. F1 gives the statistical value of optical flow where EPE is greater than $3$ pixels or the error is greater than $5\%$.

After the first stage of training, CSFlow outperforms previous methods. Consistent with the previous analysis, our accuracy on both virtual and real data is better than Flow 1D~\cite{xu2021high}, because our cost volume encodes both the global context and the fine-grained matching cost, thus showing great synthetic-to-real zero-shot generalization ability.
Compared to the current state-of-the-art method~\cite{teed2020raft}, the EPE of our method is reduced by $7\%$ and the F1 error is reduced by $5\%$.
After the second stage of training, we evaluate the model on the test set of Sinel and KITTI-2015. When only fine-tuning is performed on the target training set, the accuracy of CSFlow on Sintel-clean is improved by $12\%$ compared to RAFT, and the F1 error rate on KITTI is reduced by $4\%$. Because the strip correlation provides the global visual similarity, the foreground flow error (F1-fg) of our method is only $6.10\%$, which exceeds all previous methods. When fine-tuning on mixed data, our model still retains high performance on the test set. It is worth noting that we outperform RAFT by $16\%$ on Sintel-clean, and reduce error by $6\%$ under the F1-fg metric which considers flow outliers over foreground regions of complex street scenes. This capability is vital for many downstream autonomous driving applications such as scene parsing and visual odometry, because the foreground often includes moving vehicles and pedestrians.

\begin{table}[t]
   \centering
   \caption{\textbf{Timing, parameters, and memory requirement.}}
   \label{tab:timing}
   \centering
   \renewcommand\arraystretch{1.5}{\setlength{\tabcolsep}{8.2mm}{\begin{tabular}{ccc}%四个c代表有四列且内容居中
            \toprule%第1道横线
            Metric     & RAFT     & Ours     \\
            \midrule%第2道横线
            Parameters & 5.3M     & 5.6M     \\
            GPU Memory & 10.6GB   & 11.5GB   \\
            Timing     & 108.46ms & 112.11ms \\
            \bottomrule%第3道横线
         \end{tabular}}}
         \vspace{-1.5em} %减少表格与下文间距
\end{table}
We also verify that CSFlow keeps a low computational overhead (see Tab.~\ref{tab:timing}). The baseline method has $5.3M$ parameters, while our method has $5.6M$ parameters. We test the memory consumption of the model during the training process. The training data is Chairs, the batch size is $6$, and the image is cropped to $368{\times}496$. The model performs inferences on $100$ images with a resolution of $436{\times}1024$ on Sintel, the number of iterations of the update block is set to $12$, and the average inference time is calculated. We notice that RAFT takes $108.46ms$ on average, while CSFlow takes $112.11ms$ on average. Due to the orthogonal strip operation that efficiently encodes non-local scene cues, the computational complexity of our method is insignificant compared to the large performance improvement.

\subsection{Qualitative Analysis}
In this section, we compare the qualitative results and further explain the method we design. Fig.~\ref{fig:qualitative} visualizes the output flow of our method on the KITTI-2015 test set, and compared with RAFT.
In challenging regions with similar texture features, RAFT is limited because it only matches the local cost volume, while CSFlow benefits from the global visual similarity of orthogonal correlation coding, which can distinguish dense matching relationships on different feature regions.
For example, in the first row, our method does not confuse two similar cars that are traveling in the same direction.
The second row is another common case, where RAFT believe that white buildings are part of the street signs, but CSFlow can clearly distinguish the difference between the foreground and the background.
We conclude that CSFlow can exploit the global context to overcome issues in confusing texture-similar area, and significantly improve the estimation accuracy of the foreground optical flow in the complex street scene.

\section{CONCLUSIONS}
In this paper, we propose a novel method of encoding global visual similarity through cross strip correlation operations, and verify the effectiveness of this method through a series of quantitative experiments and qualitative analyses. Our method maintains the simplicity and high computational efficiency of the framework, which can overcome challenging texture loss and texture similar areas. It significantly improves the accuracy of the foreground optical flow estimation, and achieves state-of-the-art performance on the KITTI-2015 flow benchmark. We look forward to further exploring the non-local scene cues in flow estimation in the future.

%\addtolength{\textheight}{-12cm}

%%%%%%%%%%%%%%%%%%%%%%%%%%%%%%%%%%%%%%%%%%%%%%%%%%%%%%%%%%%%%%%%%%%%%%%%%%%%%%%%

%%%%%%%%%%%%%%%%%%%%%%%%%%%%%%%%%%%%%%%%%%%%%%%%%%%%%%%%%%%%%%%%%%%%%%%%%%%%%%%%

%%%%%%%%%%%%%%%%%%%%%%%%%%%%%%%%%%%%%%%%%%%%%%%%%%%%%%%%%%%%%%%%%%%%%%%%%%%%%%%%

%%%%%%%%%%%%%%%%%%%%%%%%%%%%%%%%%%%%%%%%%%%%%%%%%%%%%%%%%%%%%%%%%%%%%%%%%%%%%%%%

\bibliographystyle{IEEEtran}
\bibliography{ref}

\begin{thebibliography}{10}
\providecommand{\url}[1]{#1}
\csname url@samestyle\endcsname
\providecommand{\newblock}{\relax}
\providecommand{\bibinfo}[2]{#2}
\providecommand{\BIBentrySTDinterwordspacing}{\spaceskip=0pt\relax}
\providecommand{\BIBentryALTinterwordstretchfactor}{4}
\providecommand{\BIBentryALTinterwordspacing}{\spaceskip=\fontdimen2\font plus
\BIBentryALTinterwordstretchfactor\fontdimen3\font minus
  \fontdimen4\font\relax}
\providecommand{\BIBforeignlanguage}[2]{{%
\expandafter\ifx\csname l@#1\endcsname\relax
\typeout{** WARNING: IEEEtran.bst: No hyphenation pattern has been}%
\typeout{** loaded for the language `#1'. Using the pattern for}%
\typeout{** the default language instead.}%
\else
\language=\csname l@#1\endcsname
\fi
#2}}
\providecommand{\BIBdecl}{\relax}
\BIBdecl

\bibitem{horn1981determining}
B.~K. Horn and B.~G. Schunck, ``Determining optical flow,'' \emph{Artificial
  Intelligence}, 1981.

\bibitem{kondermann2016hci}
D.~Kondermann \emph{et~al.}, ``The {HCI} benchmark suite: Stereo and flow
  ground truth with uncertainties for urban autonomous driving,'' in
  \emph{CVPRW}, 2016.

\bibitem{li2019learning}
Z.~Li \emph{et~al.}, ``Learning the depths of moving people by watching frozen
  people,'' in \emph{CVPR}, 2019.

\bibitem{geiger2013vision}
A.~Geiger, P.~Lenz, C.~Stiller, and R.~Urtasun, ``Vision meets robotics: The
  {KITTI} dataset,'' \emph{The International Journal of Robotics Research},
  2013.

\bibitem{gadde2017semantic}
R.~Gadde, V.~Jampani, and P.~V. Gehler, ``Semantic video {CNNs} through
  representation warping,'' in \emph{ICCV}, 2017.

\bibitem{mustikovela2016can}
S.~K. Mustikovela, M.~Y. Yang, and C.~Rother, ``Can ground truth label
  propagation from video help semantic segmentation?'' in \emph{ECCV}, 2016.

\bibitem{zhu2017deep}
X.~Zhu, Y.~Xiong, J.~Dai, L.~Yuan, and Y.~Wei, ``Deep feature flow for video
  recognition,'' in \emph{CVPR}, 2017.

\bibitem{min2020voldor}
Z.~Min, Y.~Yang, and E.~Dunn, ``{VOLDOR:} {Visual} odometry from log-logistic
  dense optical flow residuals,'' in \emph{CVPR}, 2020.

\bibitem{teed2021droid}
Z.~Teed and J.~Deng, ``{DROID-SLAM:} {Deep} visual {SLAM} for monocular,
  stereo, and {RGB-D} cameras,'' \emph{NeurIPS}, 2021.

\bibitem{dosovitskiy2015flownet}
A.~Dosovitskiy \emph{et~al.}, ``{FlowNet:} {Learning} optical flow with
  convolutional networks,'' in \emph{ICCV}, 2015.

\bibitem{ilg2017flownet}
E.~Ilg, N.~Mayer, T.~Saikia, M.~Keuper, A.~Dosovitskiy, and T.~Brox,
  ``{FlowNet} 2.0: {Evolution} of optical flow estimation with deep networks,''
  in \emph{CVPR}, 2017.

\bibitem{hui2018liteflownet}
T.-W. Hui, X.~Tang, and C.~C. Loy, ``{LiteFlowNet:} {A} lightweight
  convolutional neural network for optical flow estimation,'' in \emph{CVPR},
  2018.

\bibitem{sun2018pwc}
D.~Sun, X.~Yang, M.-Y. Liu, and J.~Kautz, ``{PWC-net:} {CNNs} for optical flow
  using pyramid, warping, and cost volume,'' in \emph{CVPR}, 2018.

\bibitem{teed2020raft}
Z.~Teed and J.~Deng, ``{RAFT:} {Recurrent} all-pairs field transforms for
  optical flow,'' in \emph{ECCV}, 2020.

\bibitem{xu2021high}
H.~Xu, J.~Yang, J.~Cai, J.~Zhang, and X.~Tong, ``High-resolution optical flow
  from {1D} attention and correlation,'' in \emph{ICCV}, 2021.

\bibitem{wang2018non}
X.~Wang, R.~Girshick, A.~Gupta, and K.~He, ``Non-local neural networks,'' in
  \emph{CVPR}, 2018.

\bibitem{chen2017deeplab}
L.-C. Chen, G.~Papandreou, I.~Kokkinos, K.~Murphy, and A.~L. Yuille,
  ``{DeepLab:} {Semantic} image segmentation with deep convolutional nets,
  atrous convolution, and fully connected {CRFs},'' \emph{IEEE Transactions on
  Pattern Analysis and Machine Intelligence}, 2018.

\bibitem{vaswani2017attention}
A.~Vaswani \emph{et~al.}, ``Attention is all you need,'' in \emph{NeurIPS},
  2017.

\bibitem{woo2018cbam}
S.~Woo, J.~Park, J.-Y. Lee, and I.~S. Kweon, ``{CBAM:} {Convolutional} block
  attention module,'' in \emph{ECCV}, 2018.

\bibitem{cheng2016long}
J.~Cheng, L.~Dong, and M.~Lapata, ``Long short-term memory-networks for machine
  reading,'' in \emph{EMNLP}, 2016.

\bibitem{yang2019volumetric}
G.~Yang and D.~Ramanan, ``Volumetric correspondence networks for optical
  flow,'' \emph{NeurIPS}, 2019.

\bibitem{hui2020lightweight}
T.-W. Hui, X.~Tang, and C.~C. Loy, ``A lightweight optical flow {CNN} -
  {Revisiting} data fidelity and regularization,'' \emph{IEEE Transactions on
  Pattern Analysis and Machine Intelligence}, 2021.

\bibitem{revaud2015epicflow}
J.~Revaud, P.~Weinzaepfel, Z.~Harchaoui, and C.~Schmid, ``{EpicFlow:}
  {Edge-preserving} interpolation of correspondences for optical flow,'' in
  \emph{CVPR}, 2015.

\bibitem{hofinger2020improving}
M.~Hofinger, S.~R. Bul{\`o}, L.~Porzi, A.~Knapitsch, T.~Pock, and
  P.~Kontschieder, ``Improving optical flow on a pyramid level,'' in
  \emph{ECCV}, 2020.

\bibitem{hosni2012fast}
A.~Hosni, C.~Rhemann, M.~Bleyer, C.~Rother, and M.~Gelautz, ``Fast cost-volume
  filtering for visual correspondence and beyond,'' \emph{IEEE Transactions on
  Pattern Analysis and Machine Intelligence}, 2013.

\bibitem{hui2020liteflownet3}
T.-W. Hui and C.~C. Loy, ``{LiteFlowNet3:} {Resolving} correspondence ambiguity
  for more accurate optical flow estimation,'' in \emph{ECCV}, 2020.

\bibitem{lu2020devon}
Y.~Lu, J.~Valmadre, H.~Wang, J.~Kannala, M.~Harandi, and P.~Torr, ``{Devon:}
  {Deformable} volume network for learning optical flow,'' in \emph{WACV},
  2020.

\bibitem{wang2020displacement}
J.~Wang, Y.~Zhong, Y.~Dai, K.~Zhang, P.~Ji, and H.~Li, ``Displacement-invariant
  matching cost learning for accurate optical flow estimation,'' \emph{arXiv
  preprint arXiv:2010.14851}, 2020.

\bibitem{hirschmuller2007stereo}
H.~Hirschmuller, ``Stereo processing by semiglobal matching and mutual
  information,'' \emph{IEEE Transactions on Pattern Analysis and Machine
  Intelligence}, 2008.

\bibitem{butler2012naturalistic}
D.~J. Butler, J.~Wulff, G.~B. Stanley, and M.~J. Black, ``A naturalistic open
  source movie for optical flow evaluation,'' in \emph{ECCV}, 2012.

\bibitem{mayer2016large}
N.~Mayer \emph{et~al.}, ``A large dataset to train convolutional networks for
  disparity, optical flow, and scene flow estimation,'' in \emph{CVPR}, 2016.

\bibitem{loshchilov2017decoupled}
I.~Loshchilov and F.~Hutter, ``Decoupled weight decay regularization,''
  \emph{arXiv preprint arXiv:1711.05101}, 2017.

\bibitem{smith2019super}
L.~N. Smith and N.~Topin, ``Super-convergence: Very fast training of neural
  networks using large learning rates,'' in \emph{SPIE}, 2019.

\bibitem{yin2019hierarchical}
Z.~Yin, T.~Darrell, and F.~Yu, ``Hierarchical discrete distribution
  decomposition for match density estimation,'' in \emph{CVPR}, 2019.

\bibitem{zhao2020maskflownet}
S.~Zhao, Y.~Sheng, Y.~Dong, E.~I. Chang, and Y.~Xu, ``{MaskFlownet:}
  {Asymmetric} feature matching with learnable occlusion mask,'' in
  \emph{CVPR}, 2020.

\bibitem{lipson2021raft}
L.~Lipson, Z.~Teed, and J.~Deng, ``{RAFT-stereo:} {Multilevel} recurrent field
  transforms for stereo matching,'' in \emph{3DV}, 2021.

\bibitem{hur2019iterative}
J.~Hur and S.~Roth, ``Iterative residual refinement for joint optical flow and
  occlusion estimation,'' in \emph{CVPR}, 2019.

\bibitem{bar2020scopeflow}
A.~Bar-Haim and L.~Wolf, ``{ScopeFlow:} {Dynamic} scene scoping for optical
  flow,'' in \emph{CVPR}, 2020.

\bibitem{xiao2020learnable}
T.~Xiao \emph{et~al.}, ``Learnable cost volume using the {Cayley}
  representation,'' in \emph{ECCV}, 2020.

\bibitem{sun2019models}
D.~Sun, X.~Yang, M.-Y. Liu, and J.~Kautz, ``Models matter, so does training: An
  empirical study of {CNNs} for optical flow estimation,'' \emph{IEEE
  Transactions on Pattern Analysis and Machine Intelligence}, 2020.

\end{thebibliography}

\end{document}